\documentclass[letterpaper]{article} 
\usepackage{aaai23}  
\usepackage{times}  
\usepackage{helvet}  
\usepackage{courier}  
\usepackage[hyphens]{url}  
\usepackage{graphicx} 
\usepackage{natbib}  
\usepackage{caption} 
\usepackage{tikz}
\usepackage{comment}
\usepackage{amsmath,amssymb} 

\newtheorem{theorem}{Theorem}
\usepackage[accsupp]{axessibility}  

\usepackage{xcolor}
\usepackage{multirow}
\usepackage{subfigure}
\usepackage{bm}
\usepackage{booktabs}
\usepackage{subfigure}
\usepackage{graphics}

\usepackage{algorithm}
\usepackage{algorithmic}

\usepackage{newfloat}
\usepackage{listings}
\DeclareCaptionStyle{ruled}{labelfont=normalfont,labelsep=colon,strut=off} 
\floatstyle{ruled}
\newfloat{listing}{tb}{lst}{}
\floatname{listing}{Listing}

\title{Class Overwhelms: Mutual Conditional Blended-Target Domain Adaptation}

\author{
    Pengcheng Xu,\textsuperscript{\rm 1}
    Boyu Wang, \textsuperscript{\rm 1, \rm 2 *}
    Charles Ling \textsuperscript{\rm 1}
}
\affiliations{
    \textsuperscript{\rm 1} Western University, London, ON N6A 5B7, Canada\\
    \textsuperscript{\rm 2} Vector Institute, Toronto, ON M5G 1M1, Canada \\
    pxu67@uwo.ca, bwang@csd.uwo.ca, charles.ling@uwo.ca
}
\begin{document}

\maketitle

\begin{abstract}
Current methods of blended targets domain adaptation (BTDA) usually infer or consider domain label information but underemphasize hybrid categorical feature structures of targets, which yields limited performance, especially under the label distribution shift.
We demonstrate that domain labels are not directly necessary for BTDA if categorical distributions of various domains are sufficiently aligned even facing the imbalance of domains and the label distribution shift of classes.
However, we observe that the cluster assumption in BTDA does not comprehensively hold. The hybrid categorical feature space hinders the modeling of categorical distributions and the generation of reliable pseudo labels for categorical alignment.
To address these, we propose a categorical domain discriminator guided by uncertainty to explicitly model and directly align categorical distributions $P(Z|Y)$. Simultaneously, we utilize the low-level features to augment the single source features with diverse target styles to rectify the biased classifier $P(Y|Z)$ among diverse targets. Such a mutual conditional alignment of $P(Z|Y)$ and $P(Y|Z)$ forms a mutual reinforced mechanism.
Our approach outperforms the state-of-the-art in BTDA even compared with methods utilizing domain labels, especially under the label distribution shift, and in single target DA on DomainNet. Source codes are available at \url{https://github.com/Pengchengpcx/Class-overwhelms-Mutual-Conditional-Blended-Target-Domain-Adaptation}
\end{abstract}

\section{Introduction}
Deep learning suffers a serious performance drop under the distribution shift~\cite{ben2006analysis}. Unsupervised domain adaptation (UDA) is proposed to adapt a source model to a new unlabeled target domain. Most UDA research considers the adaptation from single or multiple sources to a single target (STDA). However, in reality, the target domain can be diverse and include various styles and textures, and the distribution of each class also varies from each target. These steer us to consider a practical yet challenging setting termed as \textit{blended targets domain adaptation (BTDA)}: 1)Adaptation is conducted from one single source to multiple targets. 2)Neither domain labels nor class labels are available on targets and the model should perform well on each target. 3)Label distributions of different targets can be different (label shift). In the following, we first present the analysis of BTDA, and discuss limitations of current methods due to these essential issues. Finally, we discuss that domain labels are not \textit{directly} necessary for BTDA and propose the category-oriented mutual conditional domain adaptation (MCDA), which also generalizes to common settings.

\begin{figure*}[t] 
\centering 
\includegraphics[width=0.93\textwidth]{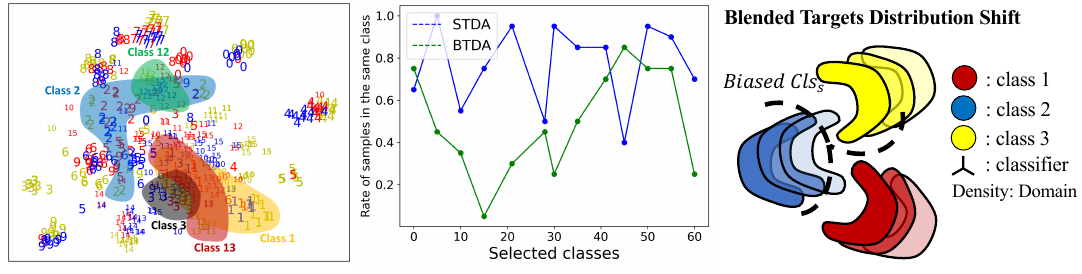} 
\caption{Left: t-SNE for hybrid categorical feature space of BTDA where features of various classes are pervasive and unstructured. The color indicates the domain and the digit indicates the class. Middle: the sample rate of the same class for each class center's K nearest neighbors. All data are collected from Office-Home (ResNet-50). Right: BTDA distribution shift where features are unstructured and the classifier is biased.} 
\label{btda_hybrid_feat} 
\end{figure*}

There are two practical issues for distributional alignment in BTDA: 1)Diverse styles and textures of blended targets. 2)Label shift of various targets. These induced our key observation that categorical feature space in BTDA is hybrid and unstructured as shown in Figure~\ref{btda_hybrid_feat}. Features of different classes in the blended targets are pervasive and do not form a well-clustered structure. To analyze it, we conduct t-SNE for feature space under BTDA in the left. Besides, we also uniformly sample and calculate K nearest neighbors (KNN) of each class center under STDA and BTDA. The result in the middle shows that the number of samples within the same class in STDA is more than that of BTDA. This indicates that the cluster structure of BTDA is not well formed compared to STDA which corresponds to the hybrid categorical feature space in t-SNE visualization.
This \textit{weakens} the cluster assumption~\cite{chapelle2005semi} that serves as the necessary condition of many adaptation methods~\cite{tachet2020domain,shu2018dirt,tang2020unsupervised,yang2021exploiting}. Further, it motivates our analytical perspective from both categorical distribution shift and biased classifier for BTDA.

Current UDA methods yield sub-optimal performance in BTDA due to these. Concretely, methods based on the \textit{covariate shift} assumption and aligning marginal distributions~\cite{tzeng2017adversarial,ganin2015unsupervised,shen2018wasserstein} are inevitable to increase the joint error of optimal hypotheses under the label shift~\cite{wu2019domain,tachet2020domain}. BTDA worsens the situation with diverse target domains and the more serious imbalance and label shift issues. Some theories further propose conditional alignment formulation to avoid the joint error issue. ~\cite{tachet2020domain,jiang2020implicit} implicitly align conditional distribution by aligning reweighted marginal distributions that still needs cluster assumption. Other methods use the target pseudo labels to model and align conditional distributions through class centroids~\cite{pan2019transferrable,tanwisuth2021prototype,singh2021clda}, task-oriented classifiers~\cite{zhang2019bridging,saito2019semi}, and the conditional discriminator~\cite{long2018conditional}. These effective STDA methods produce limited performance under the hybrid feature space in BTDA. Centroid and general adversarial methods may not model distributions well. The biased classifier and clustering labeling algorithm also generate noisy labels in this situation.

Recent multi-target domain adaptation (MTDA) methods produce impressive results by generally inferring or utilizing the domain level information and then conducting STDA methods. Some methods train separated models for each target which is not efficient in practice~\cite{saporta2021multi,isobe2021multi,nguyen2021unsupervised,saporta2021multi}. Other methods utilize graph neural networks with co-teaching~\cite{roy2021curriculum}, disentanglement methods~\cite{gholami2020unsupervised} and meta-clustering~\cite{chen2019blending}. These require domain labels and lack consideration for the imbalance and the hybrid target feature space in BTDA.

In this paper, we address two intrinsic issues of BTDA: 1)Domain labels. 2)Hybrid categorical feature space. First, our analysis shows that domain labels are not \textit{directly} necessary for BTDA only if the categorical distributions of various domains are sufficiently aligned even facing the imbalance and the label shift. However, categorical alignment requires labels. The hybrid categorical feature space in BTDA raises practical issues in modeling categorical distributions and producing reliable pseudo labels. Considering these, we design techniques to explicitly model and align categorical distributions $P(Z|Y)$ of various domains and simultaneously correct the biased classifier $P(Y|Z)$ among diverse targets to enhance pseudo labels.

Practically, this motivates two designs on $P(Z|Y)$ and $P(Y|Z)$. Firstly, for modeling and aligning $P(Z|Y)$, current methods such as prototype~\cite{pan2019transferrable,tanwisuth2021prototype} and kernel methods~\cite{wang2020rethink} inferiorly model conditional distributions of unstructured data features in the hybrid feature space  in Figure~\ref{btda_hybrid_feat}. Leveraging the distribution modeling ability of GAN~\cite{arora2017generalization,goodfellow2014generative}, we propose an uncertainty-guided categorical domain discriminator. We encode categorical distributions within the same semantic space to \textit{explicitly} model and \textit{directly} align $P(Z|Y)$ of various domains. Since the discriminator is supervised with source and noisy target labels, we adopt uncertainty to guide it to gradually learn and align categorical distributions. Secondly, to correct the biased classifier for reliable pseudo labels during adaptation, we first adopt balanced sampling on the source data and then utilize the low-level features in convolution neural networks (CNN) to augment the source features with diverse target styles to
reduce domain dependent information and balance the classifier training on target classes. Our method shows that one single labeled source can still be augmented with multiple targets to rectify the classifier during adaptation by leveraging the prior of low-level features in CNN.

In summary, our contributions are as follows: 1)We demonstrate that the adaptation can be well-achieved without domain labels in BTDA if categorical distributions are sufficiently aligned even facing the imbalance and label shift.
2)We propose the mutual conditional alignment to directly minimize conditional distributions and simultaneously correct the biased classifier.
3)Practically, to address the hybrid feature space of BTDA, we design an uncertainty-guided categorical domain discriminator to explicitly model and align categorical distributions, and utilize low-level features to mitigates the bias of classifier on blended targets.
Our method achieves the state-of-the-art in BTDA even compared with methods using domain labels, especially under the label shift, and in STDA with DomainNet.

\section{Related Works} 
\noindent\textbf{Single target UDA (STDA):} STDA is a typical setting that adapts  single or multiple sources into one target. Generally, the research includes four categories. One branch minimizes the explicit statistical distance such as Maximum Mean Discrepancy (MMD) to mitigate the domain distribution shift~\cite{long2015learning,long2017deep,officehome,tzeng2014deep,shen2018wasserstein,lee2019sliced,xu2019wasserstein,montesuma2021wasserstein}. The second branch leverages the adversarial training to implicitly minimize the domain discrepancy through GAN~\cite{ganin2015unsupervised,tzeng2017adversarial,zhang2019bridging} or entropy minimization~\cite{pan2020unsupervised,vu2019advent}. The third one utilizes the self-training with the target pseudo labels to train the source model~\cite{liu2021cycle,french2017self}. The fourth one utilizes image translation techniques to mitigate the semantic irrelevant gap~\cite{sankaranarayanan2018generate,roy2021trigan,kim2020learning,yang2020label}. However, these methods produce limited performance in BTDA. The serious imbalance and label shift issues in blended targets cause a serious incremental error of classifiers~\cite{wu2019domain}, and the hybrid target feature space also yields noisy pseudo labels and calibration issues~\cite{mei2020instance}, which deteriorates the self-training and conditional alignment methods. 

\noindent\textbf{Multi-target UDA (MTDA):} transfers the knowledge from a single source to multiple targets. MTDA is recently studied in both classification~\cite{gholami2020unsupervised,nguyen2021unsupervised,chen2019blending,roy2021curriculum,yang2020heterogeneous} and semantic segmentation ~\cite{saporta2021multi,isobe2021multi}. One common approach is to disentangle domain information from multiple targets by adversarial learning and adapt each target with a separated network~\cite{saporta2021multi,gholami2020unsupervised}. AMEAN~\cite{chen2019blending} first clusters blended targets into sub-clusters and adapts the source with each cluster. CGCT~\cite{roy2021curriculum} use graph convolution network (GCN) for feature aggregation, and use GCN classifier and source classifier for co-teaching. Differently, our method does not require any domain label and conducts BTDA in an united network which is scalable and efficient. Besides, our model considers the hybrid categorical feature space and is robust under the imbalance and label shift in BTDA.

\section{Methodology}
We present our analysis of BTDA and discuss the proposed mutual conditional domain adaptation (MCDA) framework including: explicit categorical adversarial alignment, uncertainty-guided discriminative adversarial training, and low-level feature manipulation for the classifier correction.

\noindent\textbf{Notation.} Let us denote the input-output space $\mathcal{X} \times \mathcal{Y}$ where $\mathcal{X}$ represents the image space and $\mathcal{Y}$ represents the label space. The labeled source domain is denoted as $\mathcal{S}=\{x_i^s, y_i^s \}^{|S|}_{i=1}$ and each unlabeled target is denoted as $\mathcal{T}_j=\{x_i^{t_j} \}^{|T_j|}_{i=1}$. Both source and target domains are i.i.d sampled from some distribution $P_{\mathcal{S}}(X, Y)$ and $P_{\mathcal{T}_j}(X, Y)$. For the model, we denote the feature extractor $g: \mathcal{X} \to \mathcal{Z}$ and the classifier $h: \mathcal{Z} \to \mathcal{Y}$. The error rate of the model on source $\mathcal{S}$ and target $\mathcal{T}_j$ are $\epsilon_S$ an $\epsilon_{T_j}$, and the blended target error rate is evaluated as $\epsilon_T = \frac{1}{K} \sum_{j} \epsilon_{T_j}$.

MCDA is a unified framework that adapts a single source to the blended targets such that the model performs well on each single target even facing the label distribution shift across various targets. i.e., $P_{\mathcal{S}}(Y) \neq P_{\mathcal{T}_j}(Y)$ and $ P_{\mathcal{T}_j}(Y) \neq P_{\mathcal{T}_m}(Y)$.
As proved in \cite{tachet2020domain}, minimizing marginal distribution shift of $P_{\mathcal{S}}(X)$ and $P_{\mathcal{T}}(X)$ can arbitrarily increase the target error $\epsilon_T$ due to the label shift, which fails the adaptation. The situation becomes worse in BTDA since each target can have a different $P_{\mathcal{T}_m}(Y)$. In that, we are interested to have a bound such that each term of it can be independently minimized as much as possible, and that is better irrelevant with domain labels.

\noindent\textbf{Blended error decomposition theorem.} Inspired by the generalized label shift (GLS) theorem in 
\cite{tachet2020domain}, we intend to align the conditional distributions of each class $P(Z|Y)$ within each target to the same class in the source domain on the feature space $\mathcal{Z}$. 
\begin{theorem}
For any classifier $\hat{Y}=(h\circ g)(X)$, the blended target error rate is
\begin{small}
\begin{equation}
\begin{split}
\| \epsilon_S  - \frac{1}{K} \sum_{j}^K \epsilon_{T_j} \| \leq & \frac{1}{K} \sum_{j}^K \| P_{\mathcal{S}}(Y) - P_{\mathcal{T}_j}(Y) \|_1 BER_{P_{\mathcal{S}}}(\hat{Y}\|Y) \\
& + 2(c-1)\Delta_{BTCE}(\hat{Y}).
\end{split}
\end{equation}
\label{BER}
\begin{equation}
BER_{P_{\mathcal{S}}}(\hat{Y}\|Y) = \max_{j \in [k]} P_{\mathcal{S}}(\hat{Y} \neq Y | Y = j)
\end{equation}
\begin{equation}
\begin{split}
\Delta_{BTCE}(\hat{Y}) = & \frac{1}{K} \sum_{j}^K \max_{y \neq y^\prime \in \mathcal{Y}^2}| P_{\mathcal{S}}(\hat{Y} = Y | Y=y)  \\
& - P_{\mathcal{T}_j}(\hat{Y} = Y | Y=y)|
\end{split}
\end{equation}
\end{small}
\end{theorem}
where $\| P_{\mathcal{S}}(Y) - P_{\mathcal{T}_j}(Y) \|_1$ represents the $L_1$ distance of label distributions between the source and each target and is a constant only depending on the data, $BER_{P_{\mathcal{S}}}(\hat{Y}\|Y)$ is the classification performance only related with the source domain. $\Delta_{BTCE}(\hat{Y})$ measures the conditional distribution discrepancy of each class between the source and each target. In this sense, we only need to minimize the $\Delta_{BTCE}(\hat{Y})$, which is equivalent to minimize the discrepancy between $P_{\mathcal{S}}(Z | Y=y)$ and $P_{\mathcal{T}_j}(Z | Y=y)$.

\noindent\textbf{Key differences.} First, different from \cite{tachet2020domain}, we argue that the theorem 3.3: clustering structure assumption in \cite{tachet2020domain} is a strong assumption in BTDA because each target has a different cluster structure $Z_{\mathcal{T}_j}$ in feature space under the pretrained feature extractor $g_{\mathcal{S}}$, which induces hybrid categorical feature space and different decision boundaries for different target $\mathcal{T}_j$ as illustrated in Figure~\ref{btda_hybrid_feat}. When blended together, the cluster of class $a$ in $\mathcal{T}_m$ may overlap with the cluster of class $b$ in $\mathcal{T}_n$. Consequently, the sufficient condition for GLS may not hold. Thus, calculating class ratios and aligning reweighted marginal distributions in \cite{tachet2020domain} do not induce GLS to align the semantic conditional distributions. Second, when the number of classes $ | \mathcal{Y} |$ is large, solving a quadratic problem to find class ratios require $O(|\mathcal{Y}|^3)$ time which is not efficient and accurate. We do not calculate class ratios. Finally, we do not make any assumption to satisfy GLS. Instead, we adopt the general bound in equation \ref{BER} and design model to directly minimize the conditional JS divergence $\mathcal{D}_{JS}(P_{\mathcal{S}}(Z | Y=y) \| P_{\mathcal{T}_j}(Z | Y=y)) $ to enforce it. However, aligning conditional distribution requires accurate pseudo labels. This motivates us to develop a mutual conditional alignment system to align $P(Z|Y)$ and $P(Y|Z)$ simultaneously. Besides, since we only use the class label, the domain label is unnecessary, which suits the BTDA setting. 

\begin{figure*}[t] 
\centering 
\includegraphics[width=0.95\textwidth]{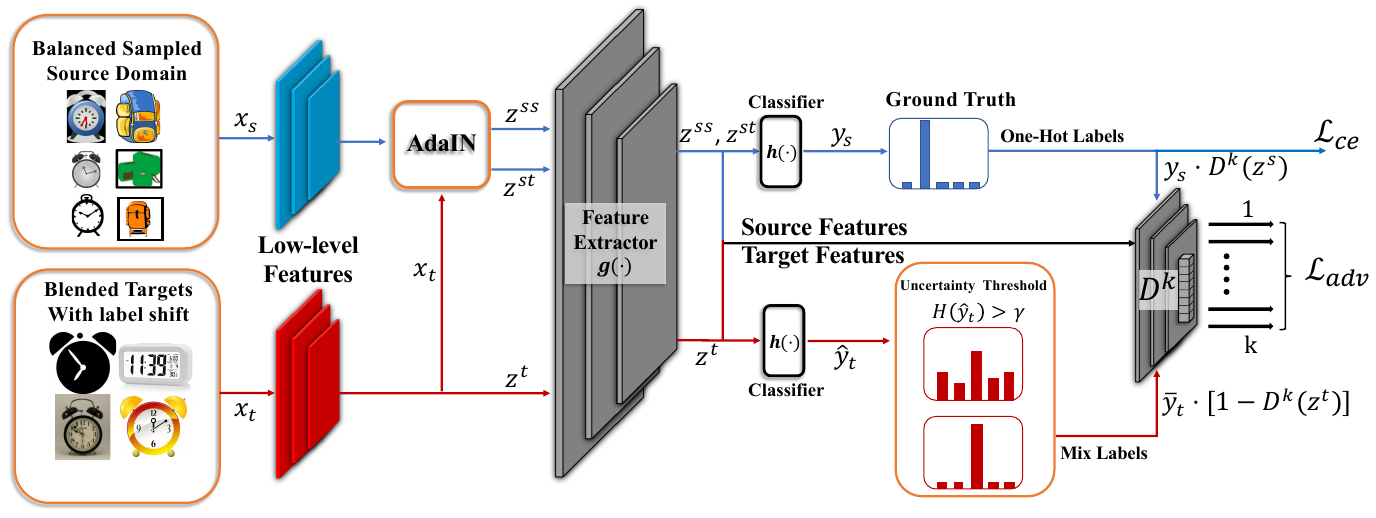} 
\caption{The framework of MCDA. The source data utilizes balanced sampling for training the categorical discriminator and is augmented with blended target styles to train the classifier. The target data is randomly sampled, and the predicted pseudo labels with low uncertainty are converted to one-hot labels to train the categorical domain discriminator.} 
\label{main_fig}
\end{figure*}

\subsection{Explicit categorical adversarial alignment}
Our motivation is to \textit{explicitly} model and \textit{directly} align the categorical JS divergence $\mathcal{D}_{JS}(P_{\mathcal{S}}(Z|Y=y) \| P_{\mathcal{T}_j}(Z|Y=y)) $ between the source and each target under the hybrid feature space. Current categorical alignment methods utilizing task-classifiers, prototypes, and conditional discriminator~\cite{zhang2019bridging,saito2019semi,long2018conditional} may not represent conditional distributions well in the hybrid categorical feature space in BTDA.

Leveraging the distribution modeling ability of GAN, we intend to encode categorical distributions of various domains into the same semantic space to explicitly model categorical distributions for optimization. Inspired by DANN~\cite{ganin2015unsupervised}, we augment the last layer of a general domain discriminator $D$ into the number of classes $k$, and each logit of such a categorical domain discriminator $D^k$ is followed by a sigmoid function to predict the probability of a feature belonging to the source or target domain conditional on the corresponding class. Each logit behaves as a single GAN to minimize the discrepancy of JS divergence of a specific class $P_\mathcal{S}(Z | Y=y)$ and $P_{\mathcal{T}_j}(Z | Y=y)$. To make each logit corresponds to one class in $D^k$, we feed one feature $g(x_i)$ into $D^k$ and get the prediction $d_i \in R^k$. Then we use the corresponding \textit{one-hot label} $y_i \in \{0,1\}^k$ to only activate the corresponding logit to compute adversarial loss by $y_i\cdot d_i$. To achieve this, we use pseudo target labels and design a strategy to make the categorical adversarial alignment and pseudo label refinement reinforce each other. Then we formulate the optimization as follows
\begin{small}
\begin{equation}
\setlength{\abovedisplayskip}{2pt}
\setlength{\belowdisplayskip}{2pt}
\begin{split}
\mathcal{L}_{adv}(g,D^k) & = \frac{1}{n_s} \sum_{i=1}^{n_s} y_i \cdot log[D^k(g(x_i^s))] + \\
    & \frac{1}{n_t} \sum_{j=1}^{n_t} \bar{y}_j \cdot log[1-D^k(g(x_j^t))],
\end{split}
\end{equation}
\end{small}
where $y_i$ represent the one-hot true labels of the source and $\bar{y}_j$ represent the mix of soft and one-hot pseudo labels of target. We discuss this in detail in the next section.

\subsection{Uncertainty guided discriminative adversarial training}
To train a discriminative categorical domain discriminator $D^k$, we require the one-hot true labels of source and blended targets. However, since the initial target labels are noisy, we design the uncertainty-guided training strategy for our categorical domain discriminator. We start with soft target labels and then gradually covert soft target labels with low uncertainty into one-hot encoding as training goes by. We use the entropy as the metric of the uncertainty of each sample, and select the samples based on a threshold $\gamma$. 
\begin{small}
\begin{equation}
\setlength{\abovedisplayskip}{2pt}
\setlength{\belowdisplayskip}{2pt}
\begin{split}
H(x_j) = - \sum_{k=1}^K \hat{y}_{j,k} \cdot log(\hat{y}_{j,k})
\end{split}
\end{equation}
\end{small}
\begin{small}
\begin{equation}
\setlength{\abovedisplayskip}{2pt}
\setlength{\belowdisplayskip}{2pt}
\bar{y}_j=\left\{
\begin{aligned}
& \{0,1\}^k,     & if H(x_j) < \gamma \\
&  \hat{y}_j,  & otherwise \\
\end{aligned}
\right.
\end{equation}
\end{small}
In the early stage, the entropy of soft pseudo labels on the target domain is large so that each digit of $D^k$ is assigned with similar probability mass. $D^k$ cannot discriminate different classes and behaves as the general discriminator $D$ in DANN since all logits share the same semantics. As the training goes, the entropy of target pseudo labels will decrease, and the labels will become more discriminative owing to the distribution alignment. At the same time, the discriminative target labels will also train $D^k$ to distinguish different categories and further align the categorical distributions, which forms a mutually reinforced process.

\subsection{Source-only balanced adversarial training}
We expect our model to be robust under the label shift across various domains such as in a case shown in Figure~\ref{btda_lmt}. Equation \ref{BER} indicates that the label shift only influences the classification error $BER_{P_{\mathcal{S}}}$ but does not influence the major distribution discrepancy $\Delta_{BTCE}$. It indicates that only if $\Delta_{BTCE}$ is small enough, the model is robust to label shift even in the blended domains. 

So, we focus on training $D^k$ since the class imbalance will lead to bias to the training of $D^k$. Hence, $D^k$ cannot distinguish different classes and align distributions biased towards the majority classes, which will ruin the categorical distribution alignment. To train a balanced $D^k$, we propose to only conduct balanced sampling for the source domain rather than on both domains as in \cite{jiang2020implicit} for two reasons: 1)We only have true labels on the source domain, balanced sampling based on the hard target pseudo labels may introduce errors and bias because initially target pseudo labels are inaccurate. Filtering out confident target pseudo labels may rule out some classes, which exacerbates the class imbalance issue. 2)With conventional double-side balanced sampling, target pseudo labels are only updated every epoch. Instead, mixed target pseudo labels can be updated online with the distribution alignment, which is more beneficial for adaptation. We demonstrate the robustness and efficiency in the Experiments section.

\subsection{Low level feature for classifier correction}
We intend to correct the biased classifier $P(Y|Z)$ from a single source to blended targets during the adaptation process. This improves the pseudo label accuracy on blended targets during the adaptation process and further facilitates the training of our categorical domain discriminator. Inspired by the research on low-level features on CNN, we utilize the low-level features of CNN that mainly represent the style and background of images to project blended target styles into the source for correcting the classifier. Denoting the low-level feature maps $z \in \mathbb{R}^{D\times H \times W}$ where $D$ represents the channel and $H,W$ represents the spatial size, leveraging the AdaIN, we have augmented features $z^{st}$ with source content and target style as below:
\begin{small}
\begin{equation}
\mu_t = \frac{1}{HW} \sum_{h=1}^H \sum_{w=1}^W z^t \hspace{2mm}
\end{equation}
\begin{equation}
\begin{split}
\sigma_t = \sqrt{\frac{1}{HW} \sum_{h=1}^H \sum_{w=1}^W (z^t - \mu_t)^2+\epsilon}
\end{split}
\end{equation}
\begin{equation}
\begin{split}
z^{st} = AdaIN(z^s, z^t) = \sigma_t (\frac{z^s - \mu_s}{\sigma_s}) + \mu_t
\end{split}
\end{equation}
\end{small}
Compared with previous image translation methods, our method does not need to generate specific images, which is efficient in practice. Besides, considering the diversity and imbalance of blended targets, our method achieves the correction on two sides: 1)Since the source is evenly resampled on class, the augmented feature $z^{st}$ with source content is balanced on semantic classes, which forms a balanced classifier for inference. 2)The augmented $z^{st}$ with diverse target styles mitigate the domain irrelevant information. This regularizes the hybrid categorical feature space in BTDA and make the cluster assumption more practical.

\noindent\textbf{Overall Objective:} Eventually, the final loss function consists of the categorical adversarial loss and the classification loss of various domains. Note that $h^{\prime}$ indicates the networks excluding the shallow layers.
\begin{small}
\begin{equation}
\setlength{\abovedisplayskip}{2pt}
\setlength{\belowdisplayskip}{2pt}
\begin{split}
\mathcal{L}_{cls}(g,h) & = \frac{1}{n_s} \sum_{i=1}^{n_s} l_{ce}(g\circ h(x_i^s), y_i^s) + \\
& \frac{1}{n_s} \sum_{i=1}^{n_s} l_{ce}(g\circ h^{\prime}(z_i^{st}), y_i^s)
\end{split}
\end{equation}
\begin{equation}
\setlength{\abovedisplayskip}{2pt}
\setlength{\belowdisplayskip}{2pt}
\begin{aligned}
\min_{g,h} \max_D \mathcal{L} = \mathcal{L}_{cls}(g,h) + \mathcal{L}_{adv}(g,D^k)
\end{aligned}
\end{equation}
\end{small}

\section{Experiments}
\begin{figure}[!htbp]
    \centering
    \includegraphics[width=1.6in]{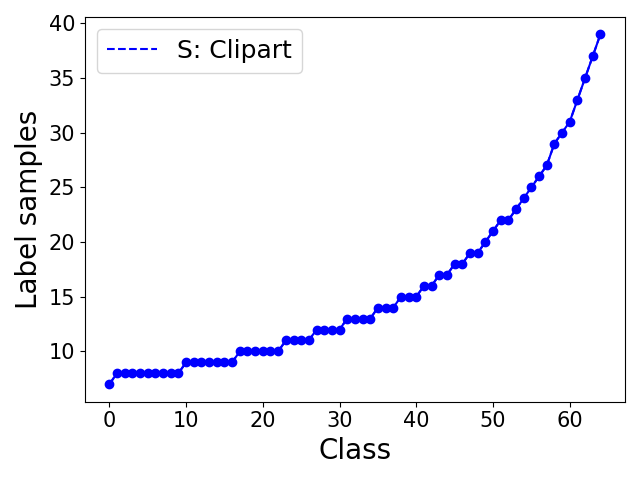}
    \includegraphics[width=1.6in]{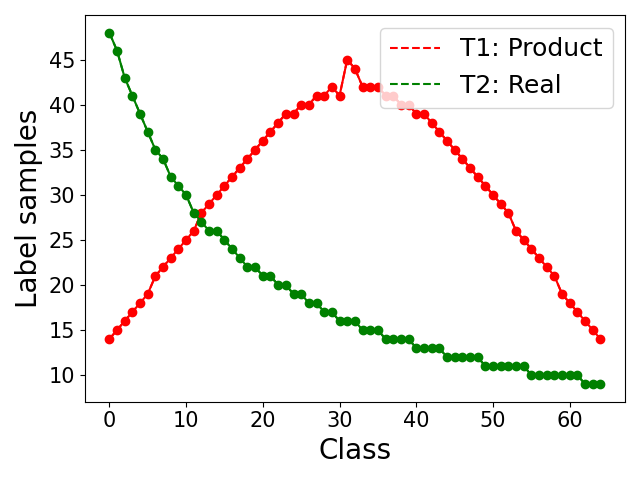}
\caption{Label distribution shift of Office-Home-LMT.}
\label{btda_lmt}
\end{figure}

\noindent\textbf{Datasets.} We evaluate our method based on standard BTDA tasks~\cite{chen2019blending,roy2021curriculum}: Office-31~\cite{office31}, Office-Home~\cite{officehome}, DomainNet~\cite{domainnet}, and a specialized dataset Office-Home-LMT for label shift in BTDA. Similar to Office-Home-RS-UT~\cite{jiang2020implicit}, we use \textbf{Cl}, \textbf{Pr} and \textbf{Rw} to resample two reverse long-tailed distributions and one Gaussian distributions for each of them for BTDA with label shift. For evaluation, we use one domain as the source and the rest as blended targets. The performance is evaluated as the mean accuracy of all target domains. We show a concrete example of \textbf{Ar} as the source in the label shift setting in Figure~\ref{btda_lmt}.

\begin{table*}[t]
\small
\centering
\setlength{\tabcolsep}{1.7mm}{
\begin{tabular}{@{}llcccccccccllccccccc@{}}
\toprule
\multicolumn{2}{l}{\multirow{2}{*}{Methods}} & \multicolumn{4}{c|}{\textbf{Office-31}}                                                                 & \multicolumn{5}{c|}{\textbf{Office-Home}}                                                                                      & \multicolumn{2}{l}{\multirow{2}{*}{Methods}} & \multicolumn{7}{c}{\textbf{DomainNet}}                                                                                                                      \\ \cmidrule(lr){3-11} \cmidrule(l){14-20} 
\multicolumn{2}{l}{}                         & A                    & D                    & W                    & \multicolumn{1}{c|}{Avg.}          & Ar                   & Cl                   & Pr                   & Rl                   & \multicolumn{1}{c|}{Avg.}          & \multicolumn{2}{l}{}                         & Cli                  & Inf                  & Pai                  & Qui                  & Rea                  & Ske                  & Avg.                 \\ \midrule
\multicolumn{2}{l}{Source}                   & 68.6                 & 70.0                 & 66.5                 & \multicolumn{1}{c|}{68.4}          & 47.6                 & 42.6                 & 44.2                 & 51.3                 & \multicolumn{1}{c|}{46.4}          & \multicolumn{2}{l}{Source}                   & 25.6                 & 16.8                 & 25.8                 & 9.2                  & 20.6                 & 22.3                 & 20.1                 \\ \midrule
\multicolumn{2}{l}{DAN}                      & 79.5                 & 80.3                 & 81.2                 & \multicolumn{1}{c|}{80.4}          & 55.6                 & 56.6                 & 48.5                 & 56.7                 & \multicolumn{1}{c|}{54.4}          & \multicolumn{2}{l}{SE}                       & 21.3                 & 8.5                  & 14.5                 & 13.8                 & 16.0                 & 19.7                 & 15.6                 \\
\multicolumn{2}{l}{DANN }                     & 80.8                 & 82.5                 & 83.2                 & \multicolumn{1}{c|}{82.2}          & 58.4                 & 58.1                 & 52.9                 & 62.1                 & \multicolumn{1}{c|}{57.9}          & \multicolumn{2}{l}{MCD}                      & 25.1                 & 19.1                 & 27.0                 & 10.4                 & 20.2                 & 22.5                 & 20.7                 \\
\multicolumn{2}{l}{CDAN}                     & 93.6                 & 80.5                 & 81.3                 & \multicolumn{1}{c|}{85.1}          & 59.5                 & 61.0                 & 54.7                 & 62.9                 & \multicolumn{1}{c|}{59.5}          & \multicolumn{2}{l}{CDAN}                     & 31.6                 & 27.1                 & 31.8                 & 12.5                 & 33.2                 & 35.8                 & 28.7                 \\
\multicolumn{2}{l}{JAN}                       & 84.2                 & 74.4                 & 72.0                 & \multicolumn{1}{c|}{76.9}          & 58.3                 & 60.5                 & 52.2                 & 57.5                 & \multicolumn{1}{c|}{57.1}          & \multicolumn{2}{l}{DADA }                     & 26.4                 & 20.0                 & 26.5                 & 12.9                 & 20.7                 & 22.8                 & 21.5                 \\
\multicolumn{2}{l}{AMEAN}                    & 90.1                 & 77.0                 & 73.4                 & \multicolumn{1}{c|}{80.2}          & 64.3                 & 65.5                 & 59.5                 & 66.7                 & \multicolumn{1}{c|}{64.0}          & \multicolumn{2}{l}{MCC}                      & 33.6                 & 30.0                 & 32.4                 & 13.5                 & 28.0                 & 35.3                 & 28.8                 \\
\multicolumn{2}{l}{CGCT}                     & \textbf{93.9}        & 85.1                 & 85.6                 & \multicolumn{1}{c|}{88.2}          & 67.4                 & 68.1                 & 61.6                 & 68.7                 & \multicolumn{1}{c|}{66.5}          & \multicolumn{2}{l}{CGCT}                     & 36.1                 & 33.3                 & 35.0                 & 10.0                 & \textbf{39.6}        & 39.7                 & 32.3                 \\ \midrule
\multicolumn{2}{l}{Ours}                     & 92.4                 & \textbf{87.7}        & \textbf{88.8}        & \multicolumn{1}{c|}{\textbf{89.6}} & \textbf{71.7}        & \textbf{72.8}        & \textbf{68.0}        & \textbf{71.7}        & \multicolumn{1}{c|}{\textbf{71.1}} & \multicolumn{2}{l}{Ours}                     & \textbf{37.5}        & \textbf{37.3}        & \textbf{36.6}        & \textbf{17.8}        & 36.1                 & \textbf{41.4}        & \textbf{34.5}        \\ \midrule 
\multicolumn{2}{l}{MTDA$^{\dag}$}     & 87.9     & 83.7     & 84.0     & \multicolumn{1}{c|}{85.2}    & 64.6      & 66.4    & 59.2    & 67.1     & \multicolumn{1}{c|}{64.3}     & \multicolumn{2}{l}{-}       & -        & -      & -      & -       & -     & -      & -       \\
\multicolumn{2}{l}{DCL$^{\dag}$}    & 92.6    & 82.5    & 84.7  & \multicolumn{1}{c|}{86.6}     & 63.0    & 63.0     & 60.0     & 67.0    & \multicolumn{1}{c|}{64.1}          & \multicolumn{2}{l}{DCL$^{\dag}$}        & 35.1        & 31.4        & 37.0        & 20.5     & 35.4     & 41.0        & 33.4          \\
\multicolumn{2}{l}{DCGCT$^{\dag}$}    & 93.4     & 86.0    & 87.1    & \multicolumn{1}{c|}{88.8}   & 70.5      & 71.6         & 66.0         & 71.2         & \multicolumn{1}{c|}{69.8}          & \multicolumn{2}{l}{DCGCT$^{\dag}$}      & 37.0                 & 32.2                 & 37.3                 & 19.3                 & 39.8                 & 40.8                 & 34.4                 \\ \midrule
                      &                      & \multicolumn{1}{l}{} & \multicolumn{1}{l}{} & \multicolumn{1}{l}{} & \multicolumn{1}{l}{}               & \multicolumn{1}{l}{} & \multicolumn{1}{l}{} & \multicolumn{1}{l}{} & \multicolumn{1}{l}{} & \multicolumn{1}{l}{}               &                       &                      & \multicolumn{1}{l}{} & \multicolumn{1}{l}{} & \multicolumn{1}{l}{} & \multicolumn{1}{l}{} & \multicolumn{1}{l}{} & \multicolumn{1}{l}{} & \multicolumn{1}{l}{} \\
                      &                      & \multicolumn{1}{l}{} & \multicolumn{1}{l}{} & \multicolumn{1}{l}{} & \multicolumn{1}{l}{}               & \multicolumn{1}{l}{} & \multicolumn{1}{l}{} & \multicolumn{1}{l}{} & \multicolumn{1}{l}{} & \multicolumn{1}{l}{}               &                       &                      & \multicolumn{1}{l}{} & \multicolumn{1}{l}{} & \multicolumn{1}{l}{} & \multicolumn{1}{l}{} & \multicolumn{1}{l}{} & \multicolumn{1}{l}{} & \multicolumn{1}{l}{}
\end{tabular}
}
\vspace{-0.8cm}
\caption{Accurary ($\%$) of BTDA on Office-31, Office-Home (ResNet-50), and  DomainNet (ResNet-101). Best results in Bold. Each domain represents the source and the rest domains are blended as the target. The accuracy is the mean of accuracies of all domains in the blended target. \dag indicates methods using domain labels.}
\label{btda_all}
\end{table*}

\noindent\textbf{Baselines and implementations.}
We compare our method with previous state-of-the-arts in standard BTDA and that with label shift, i.e., MTDA~\cite{nguyen2021unsupervised}, CGCT~\cite{roy2021curriculum}, MDDIA~\cite{jiang2020implicit}, CST~\cite{liu2021cycle}, and SENTRY~\cite{prabhu2021sentry}. For comparison with BTDA under label shift, we also combine the sampling strategies in MDDIA with CGCT. The detailed summary of comparison methods is in the supplementary. We followed the implementations in ~\cite{dalib}. For all datasets, we use SGD optimizer with learning rates $\eta_0 = 0.01$, $\alpha = 10$, and $\beta = 0.75$. We set the uncertainty threshold $\ \gamma = 0.05$ for all datasets. Since CST and SENTRY use AutoAugment for data augmentation, we set the number of transformations $N=1$ and the transform severity $M=2.0$ in AutoAugment~\cite{lim2019fast} for a fair comparison.

\begin{table}[t]
\centering
\small
\setlength{\tabcolsep}{3mm}{
\begin{tabular}{@{}lcccc@{}}
\toprule
\textbf{Methods} & Clipart & Product & Real & Avg.  \\ \midrule
Source  & 42.3    & 47.6    & 50.3 & 46.7 \\ \midrule
BSP     & 51.5    & 52.9    & 57.4 & 54.0 \\
CDAN    & 50.5    & 53.2    & 56.3 & 53.3 \\ 
DAN    & 51.0    & 49.2    & 56.8 & 52.3 \\
JAN    & 51.4    & 50.1    & 57.0 & 53.2 \\
DANN    & 46.6    & 50.4    & 53.3 & 50.1 \\
ADDA    & 45.0    & 49.7  & 52.8 & 49.2 \\
MCD    & 40.2    & 48.6    & 52.3 & 47.0 \\
MDD     & 43.7    & 56.0    & 57.8 & 52.5 \\
MDDIA   & 61.9    & 58.2    & 63.2 & 61.1  \\
CGCT    & 53.7    & 51.5    & 52.0 & 52.4   \\ 
CGCT+bal  & 57.1  & 53.0 & 56.8  & 55.7  \\
Ours    & \textbf{68.0}    & \textbf{62.3}    & \textbf{67.5} & \textbf{65.9} \\ \midrule
CST(aug)     & 58.3    & 57.4    & 63.4 & 59.7 \\
SENTRY(aug)  & 65.6    & 63.5    & 65.9 & 65.0  \\
Ours(aug)   & \textbf{69.1}  & \textbf{66.2}    & \textbf{68.9}    & \textbf{68.1}     \\ \midrule
Ours(oracle)   & 98.9  & 98.3    & 98.2    & 98.5  \\
S+T &99.7   & 99.8    & 99.8 &99.8 \\ 
\bottomrule
\end{tabular}
}
\caption{Accurary ($\%$) of Blended-Office-Home-LMT (ResNet-50). \textit{aug}: using 1 extra augmented data with RandAug~\cite{cubuk2020randaugment}. \textit{bal}: using balanced sampling. \textit{oracle}: discriminator trained with true source and target labels. \textit{S+T}: supervised learning on source and target.}
\label{btlds}
\end{table}


\noindent\textbf{Standard BTDA.} We summarize the standard BTDA in Table~\ref{btda_all}. Our method outperforms comparison methods with a clear margin. Concretely, our method outperforms the latest BTDA methods (e.g., AMEAN and CGCT) by 1.4\% on Office-31, 4.6\% on Office-Home, and 2.2\% on DomainNet. Moreover, even compared with methods utilizing ground truth domain labels, our method can still outperform them by 0.8\% on Office-31 and 1.3\% on Office-Home. The results validate our argument that categorical distribution alignment overwhelms in BTDA, and echos the theoretical intuition from equation~\ref{BER} that proper domain alignment is achievable even without domain labels in BTDA.

\noindent\textbf{BTDA with label shift.} We analyzed the essential label shift influence on BTDA and summarized results of the specialized Office-Home-LMT in Table~\ref{btlds}. The label distribution of each domain is different from each other. Since CST and SENTRY essentially require extra augmented data, we add 1 augmented data for each sample and evaluate CST, SENTRY, and ours in the same setting (\textit{aug}). Our method outperforms the label shift UDA method MDDIA by 4.8\% and SENTRY by 3.1\%. Compared with latest BTDA method CGCT, we get an improvement of more than 12\%. We also equip CGCT with balanced sampling strategy in MDDIA (i.e., CGCT+bal) whose result is still inferior to ours.

The result first demonstrates the label shift in BTDA seriously impedes the adaptation, especially for the marginal alignment methods. Second, it validates our proposition in theorem 1 that if categorical distribution $\Delta_{BTCE}(Y)$ can be properly minimized, the label shift only reweights the classification error $BER_{P_{\mathcal{S}}}(\hat{Y}\|Y)$, which is relatively small. Besides, our method does not require balanced sampling on target pseudo labels for every epoch, which can be trained and updated online. The extra data augmentation (e.g., RandAug) is not essentially necessary in the algorithm design.

\noindent\textbf{STDA.} We also validate the generalization ability of our method in STDA (i.e., Office-Home and DomainNet). We compare our method with SRDC~\cite{tang2020unsupervised} which considers cluster structures in STDA, and MDDIA~\cite{jiang2020implicit} which uses balanced sampling on both source and target domains. Our method achieve 72.4\% on Office-Home, 35.2\% on DomainNet which outperforms previous state-of-the-art method MDD+SCDA~\cite{li2021semantic} by 1.9\%. Please refer to supplementary for details.

\begin{table}[!htbp]
\centering
\small
\setlength{\tabcolsep}{0.5mm}{
\begin{tabular}{@{}lll|cccccccccc@{}}
\toprule
 \multicolumn{1}{l}{\multirow{2}{*}{mix}} & \multicolumn{1}{l}{\multirow{2}{*}{bal}} & \multicolumn{1}{l|}{\multirow{2}{*}{flip}} & \multicolumn{5}{c}{\textbf{Office-Home}}                                                                                                  &  & \multicolumn{4}{c}{\textbf{DomainNet}}                                                                    \\ \cmidrule(l){4-13} 
\multicolumn{1}{c}{}                     & \multicolumn{1}{c}{}                          & \multicolumn{1}{c|}{}                      & \multicolumn{1}{c}{Art} & \multicolumn{1}{c}{Clip} & \multicolumn{1}{c}{Prod} & \multicolumn{1}{c}{Real} & \multicolumn{1}{c}{Avg.} &  & \multicolumn{1}{c}{Real} & \multicolumn{1}{c}{Info} & \multicolumn{1}{c}{Pain} & \multicolumn{1}{c}{Avg.} \\ \midrule
\checkmark &            &   &66.0  &67.3 &64.5 &70.9 &67.2  & &32.0 &31.0 &33.4 &32.1   \\
& \checkmark &  &70.6   &72.5    &67.4    &69.9    &70.1  &    &34.3     &33.5     &34.9   &34.2 \\ 
& & \checkmark  &65.0   &66.5   &65.6    &71.0    &67.0     &  &34.6    &30.7    &32.8   &32.7  \\
\checkmark & \checkmark & &70.2 &\textbf{73.2} &67.0 &70.6 &70.3 &  &35.7  &35.7 &36.1  &35.8    \\
\checkmark & \checkmark & \checkmark  &\textbf{71.7}   &72.8   &\textbf{68.0}   &\textbf{71.7}  &\textbf{71.1}   &   &\textbf{36.1}  &\textbf{37.3}  &\textbf{36.6}  &\textbf{36.7}  \\ \bottomrule

\end{tabular}
}
\caption{Ablations on Office-Home and the selected three domains on DomainNet. \textit{mix}: mix labeling strategy; \textit{bal}: balanced sampling strategy; \textit{flip}: low-level features.}
\label{ablation}
\end{table}

\section{Ablation and analysis}
We present ablations of MCDA in Table~\ref{ablation} on Office-Home and three domains of DomainNet. Each proposed module contributes to the improvement of the final performance.

\begin{figure}[t]
\centering
    \label{fig:subfig:sourceonly_tsne} 
    \includegraphics[width=0.23\textwidth]{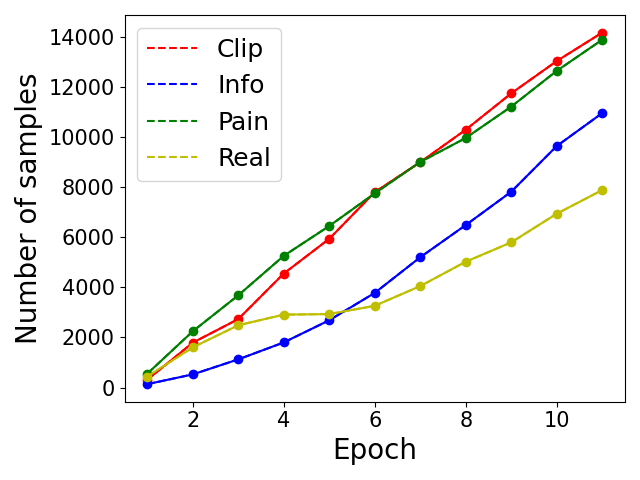}
    \label{fig:subfig:mcda_tsne} 
    \includegraphics[width=0.23\textwidth]{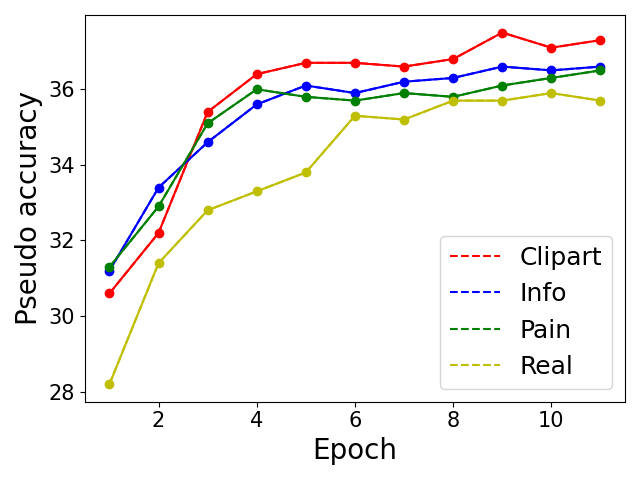}

\caption{Samples below uncertainty threshold and pseudo label accuracy during training process on DomainNet.} 
\label{pse_acc}
\end{figure}

\noindent\textbf{Effectiveness of uncertainty for guiding adversarial training.} To validate the uncertainty and mixed labels to train a categorical discriminator $D^k$ that mutually reinforces with target pseudo labels. We show the number of filtered samples below the uncertainty threshold and the corresponding pseudo label accuracy in Figure~\ref{pse_acc}. The results of DomainNet show that during the training, the uncertainty of samples gradually decreases, and more samples pass the threshold. Meanwhile, the accuracy of pseudo labels increases, which justifies our motivation.

\begin{figure}[t]
\centering

\label{fig:subfig:onefunction} 
\includegraphics[width=0.22\textwidth]{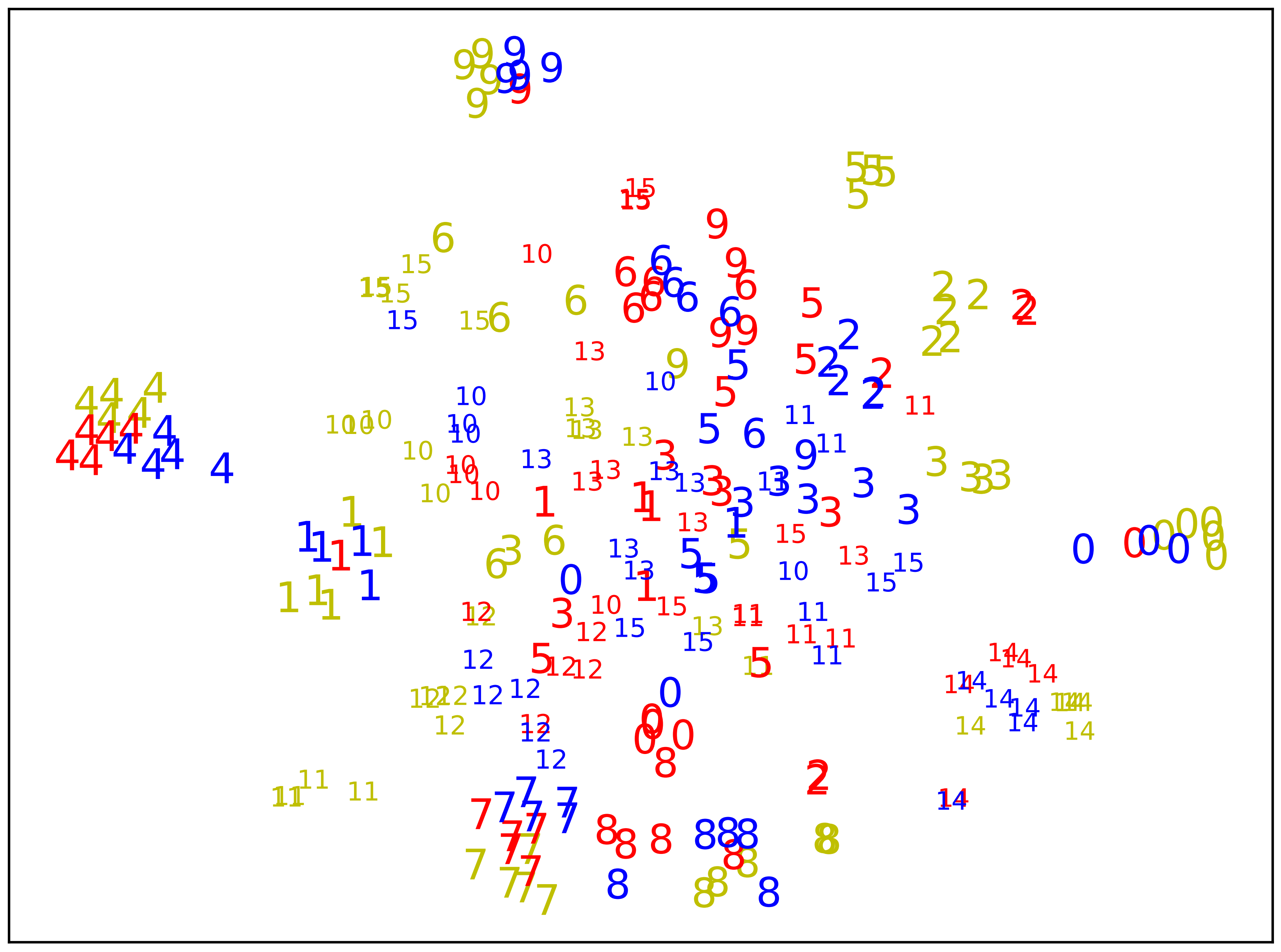}
\label{fig:subfig:twofunction} 
\includegraphics[width=0.22\textwidth]{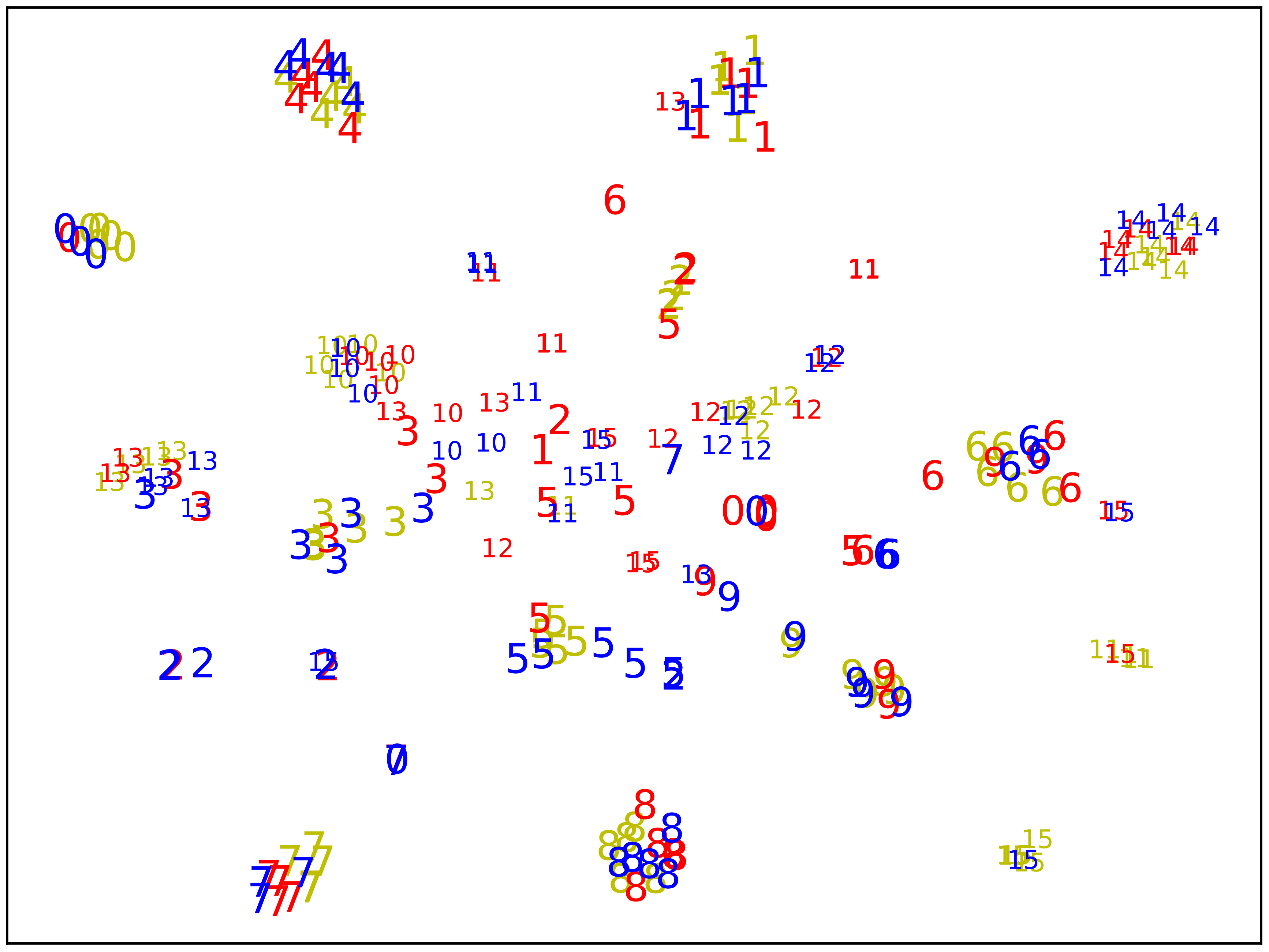}

\caption{t-SNE feature visualization on Office-Home-LMT with \textit{Clipart} as the source. 15 classes are sampled for conciseness. Color represents the domain and the digit represents the class. \textit{left}: sourceonly, \textit{right}: MCDA.} 
\label{t-sne}
\end{figure}

\begin{figure}[t]
\centering

\includegraphics[width=0.45\textwidth]{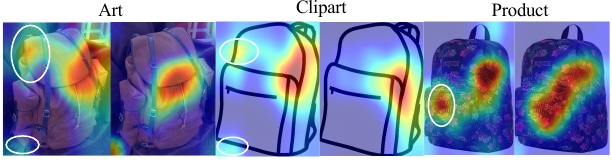}

\caption{CAM feature response maps on Office-Home-LMT. The left pervasive one is from sourceonly model while the right class-discriminative one is from MCDA.}
\label{cam} 
\end{figure}

\noindent\textbf{Robustness of uncertainty threshold.} We validate the robustness of our model under various uncertainty thresholds $\lambda$ in Table~\ref{thre} for both standard BTDA and BTDA with label shift in Table~\ref{thre}. For main experiments in Table~\ref{btda_all} and~\ref{btlds}, we set $\lambda$ as 0.05. The performance is stable when $\lambda$ is selected from 0.03 to 0.07. The results of Art on Office-Home have a fluctuation range of 0.4\%, and results of Clipart on Office-Home-LMT have a fluctuation range of 1.5\%. These demonstrate the stability and generality of our model for the standard BTDA and label shift setting.

\noindent\textbf{Verification of error theorem under label shift.} We verify our error theorem in equation 1 that if conditional distributions $\Delta_{BTCE}(\hat{Y})$ is sufficiently minimized, the model is robust under label shift in BTDA since the reweighted source error $\| P_{\mathcal{S}}(Y) - P_{\mathcal{T}_j}(Y) \|_1  BER_{P_{\mathcal{S}}}(\hat{Y}\|Y)$ is relatively small. In Table~\ref{btlds}, the \textit{oracle} are results when the discriminator is trained with true labels of source and target but the classifier is trained only with source labels. In this case, the categorical discriminator is trained to minimize categorical distributions as much as possible under ground truth supervision. The \textit{S+T} are results where the classifier is trained with both source and target labels. These two results approximate to each other which verifies the theorem.

\begin{table}[!htbp]
\centering
\setlength{\tabcolsep}{1.5mm}{
\scalebox{1.0}{
\begin{tabular}{@{}lcccccc@{}}
\toprule
\textbf{Thresholds} & 0.01 & 0.03 & 0.05 & 0.07 & 0.09 \\ \midrule
Art      & 71.9    & 71.9    & 71.7    & 72.3  & 71.8   \\ 
Clipart  & 66.5    & 67.4    & 68.0    & 66.9  & 66.4 \\
\bottomrule
\end{tabular}
}
}
\caption{Accuracy ($\%$) of different uncertainty thresholds for \textit{Art} in Office-Home and \textit{Clipart} in Office-Home-LMT for BTDA (ResNet-50).}

\label{thre}
\end{table}

\noindent\textbf{Essential of domain labels.} Our theoretical formulation in equation 1 does not require domain labels to minimize the target error rate in BTDA. The bound is mainly related to the categorical distribution constraint. 
In comparison with methods (i.e., CGCT~\cite{roy2021curriculum} and DCL~\cite{nguyen2021unsupervised}) using domain labels $\dag$ in Table~\ref{btda_all}, our method outperforms previous state-of-the-arts by 0.8\% on Office-31, 1.3\% on Office-Home, and 0.1\% on DomainNet even without domain labels.
This validates our proposition that adaptation can be done without domain labels in BTDA if categorical distributions is sufficiently aligned.

\noindent\textbf{Feature visualization.} To show our method learns a regular and meaningful categorical feature space, we visualize features of last convolution layer with t-SNE in Figure~\ref{t-sne} and CAM~\cite{selvaraju2017grad} in Figure~\ref{cam}. The t-SNE visualization further shows that the sourceonly model generates a hybrid feature space while MCDA produces a more class-discriminative feature space, which corroborates our observation on the cluster assumption and categorical alignment on BTDA. The CAM results shows feature response maps of the sourceonly model is pervasive while those of MCDA are more category-discriminative. This validates that MCDA make the classifier learns more task-relevant features and achieve better categorical alignment. More visualization results are discussed in the supplementary. 

\section{Conclusion}
In this paper, we demonstrate that adaptation can be well achieved without domain labels for BTDA only if the categorical distributions are sufficiently aligned, even facing the cross-domain label shift. Following this, we present the mutual conditional domain adaptation framework. We explore the uncertainty-guided mechanism and source-only balanced sampling strategy to train a categorical domain discriminator for efficiently modeling categorical distributions in BTDA. And we explore low-level features to correct the biased classifier. Extensive experimental results demonstrate the state-of-the-art performance of the framework in single target DA and BTDA tasks under label shift.

\section{Acknowledgement}
We appreciate constructive feedback from anonymous reviewers and meta-reviewers. This work is supported by Natural Sciences and Engineering Research Council of Canada (NSERC), Discovery Grants program.

\bibliography{egbib}
\bibliographystyle{aaai23}

\end{document}